\def\year{2021}\relax
\documentclass[letterpaper]{article} 
\usepackage{aaai21}  
\usepackage{times}  
\usepackage{helvet} 
\usepackage{courier}  
\usepackage[hyphens]{url}  
\usepackage{graphicx} 
\usepackage{natbib}  
\usepackage{caption} 
\usepackage{epsfig}
\usepackage{amsmath}
\usepackage{dsfont}
\usepackage{chngpage}
\usepackage{subcaption}
\usepackage[T1]{fontenc}

\nocopyright

\title{Robust Federated Learning with Noisy Labels}
\author{

    Seunghan Yang, Hyoungseob Park, Junyoung Byun, Changick Kim
    \\
}
\affiliations{

    KAIST, South Korea\\
    \{seunghan, hyoungseob, bjyoung, changick\}@kaist.ac.kr

}

\begin{document}

\maketitle

\begin{abstract}
Federated learning is a paradigm that enables local devices to jointly train a server model while keeping the data decentralized and private.
In federated learning, since local data are collected by clients, it is hardly guaranteed that the data are correctly annotated.
Although a lot of studies have been conducted to train the networks robust to these noisy data in a centralized setting, these algorithms still suffer from noisy labels in federated learning. 
Compared to the centralized setting, clients' data can have different noise distributions due to variations in their labeling systems or background knowledge of users.
As a result, local models form inconsistent decision boundaries and their weights severely diverge from each other, which are serious problems in federated learning.
To solve these problems, we introduce a novel federated learning scheme that the server cooperates with local models to maintain consistent decision boundaries by interchanging class-wise centroids.
These centroids are central features of local data on each device, which are aligned by the server every communication round.
Updating local models with the aligned centroids helps to form consistent decision boundaries among local models, although the noise distributions in clients' data are different from each other.
To improve local model performance, we introduce a novel approach to select confident samples that are used for updating the model with given labels. Furthermore, we propose a global-guided pseudo-labeling method to update labels of unconfident samples by exploiting the global model.
Our experimental results on the noisy CIFAR-10 dataset and the Clothing1M dataset show that our approach is noticeably effective in federated learning with noisy labels.
\end{abstract}

\section{Introduction}
Modern edge devices such as smart phones have been able to access an abundant amount of data, which is suitable for training deep learning models. Since each client device should transmit its local data to the central server for conventional centralized learning, it can lead to serious data privacy issues. To address these problems, federated learning has been actively studied to shift a learning environment from the central server to each edge device. In detail, federated learning allows a server model to be trained on each client's private data without transmitting raw data to the server.

\begin{figure}
\hspace{5.5mm}
\epsfig{file=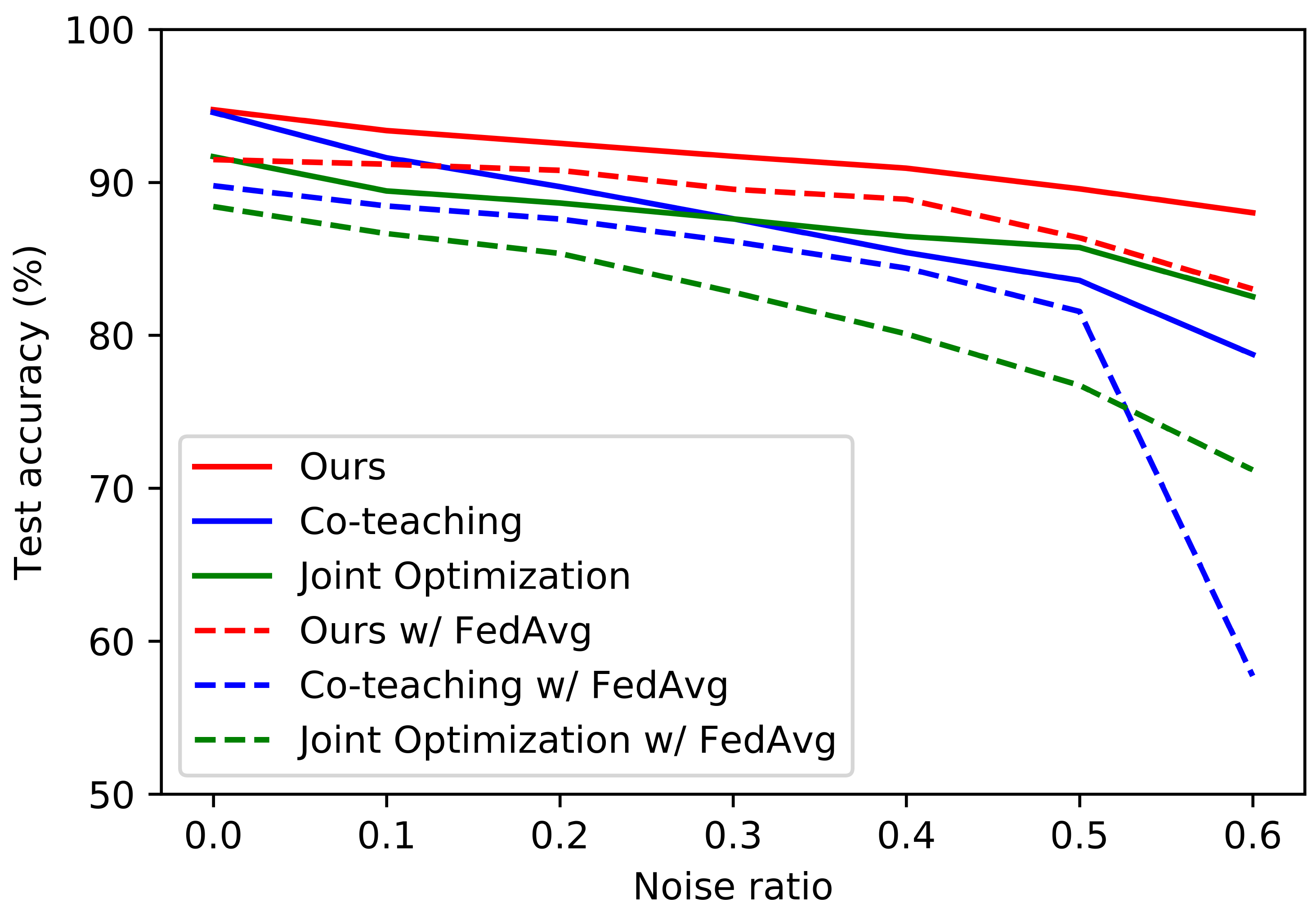,width=0.8\linewidth}
   \caption{Test accuracy on the CIFAR-10 dataset at various noise ratios in the centralized setting (solid line) and the federated setting (dotted line). For federated learning with noisy labels, we distribute noisy data to clients in an i.i.d. fashion. Co-teaching \cite{han2018co} and Joint Optimization \cite{tanaka2018joint} are novel methods for the centralized setting, but these algorithms combined with FedAvg \cite{mcmahan2016communication} suffer from performance degradation in the federated setting. Best viewed in color.}
\label{fig1}
\end{figure}

The federated learning paradigm consists of two stages: 1) In the beginning of each round, a server broadcasts the server model to selected clients, and these clients train models on their own data for multiple iterations. 2) After the clients train their models, the server aggregates the clients' model parameters. The above process iterates until the global model converges. 
In FedAvg \cite{mcmahan2016communication}, model parameters of clients are aggregated in an element-wise manner with coefficients, which are proportional to the local dataset size. The global model effectively converges by FedAvg, especially when the local dataset follows an i.i.d. distribution.
Many studies have been conducted to apply it to practical applications, {\it e.g.}, dealing with non-i.i.d. data \cite{li2018federated, zhao2018federated, shoham2019overcoming, Wang2020Federated, Li2020On}, noisy communication \cite{9026922}, domain adaptation \cite{Peng2020Federated}, fair resource allocation \cite{Li2020Fair}, and continual learning \cite{yoon2020federated}.

Although the above studies try to solve practical application issues related to preserving privacy, there are still remaining problems when local devices are used for training neural networks. 
In practice, all local data should be annotated by alternative labeling techniques such as exploiting machine-generated labels \cite{kuznetsova2018open} due to privacy issues.
These labels are inevitably corrupted unless the labeling techniques of all clients are accurate.
Similarly, in the centralized setting, robust learning with noisy labels has attracted attention due to its applicability for realistic situations, and various algorithms have been proposed to train models accurately in the presence of noise. Recent algorithms have tried to minimize the effect of the noisy labels by sampling reliable data \cite{han2018co, wei2020combating, huang2019o2u, guo2018curriculumnet}, updating labels \cite{tanaka2018joint, yi2019probabilistic}, or estimating labels from matched prototypes \cite{han2019deep, lee2018cleannet}. These approaches have evolved into training the model with noisy labels successfully.


The aforementioned approaches for dealing with noisy labels suffer from performance degradation in the federated setting, as illustrated in Fig. \ref{fig1}. Unlike centralized learning, noise distributions in clients' data can be different from each other due to the discrepancy between their labeling systems or background knowledge.
As a result, local models form inconsistent decision boundaries and their weights severely diverge from each other, {\it i.e.,} weight divergence.
This causes aggregation difficulties of local models, which are a serious problem in federated learning \cite{li2018federated, chen2020fedmax, lim2020federated}.

\begin{figure*}
\begin{center}
\hspace{10mm}
\begin{minipage}{0.35\linewidth}
\epsfig{file=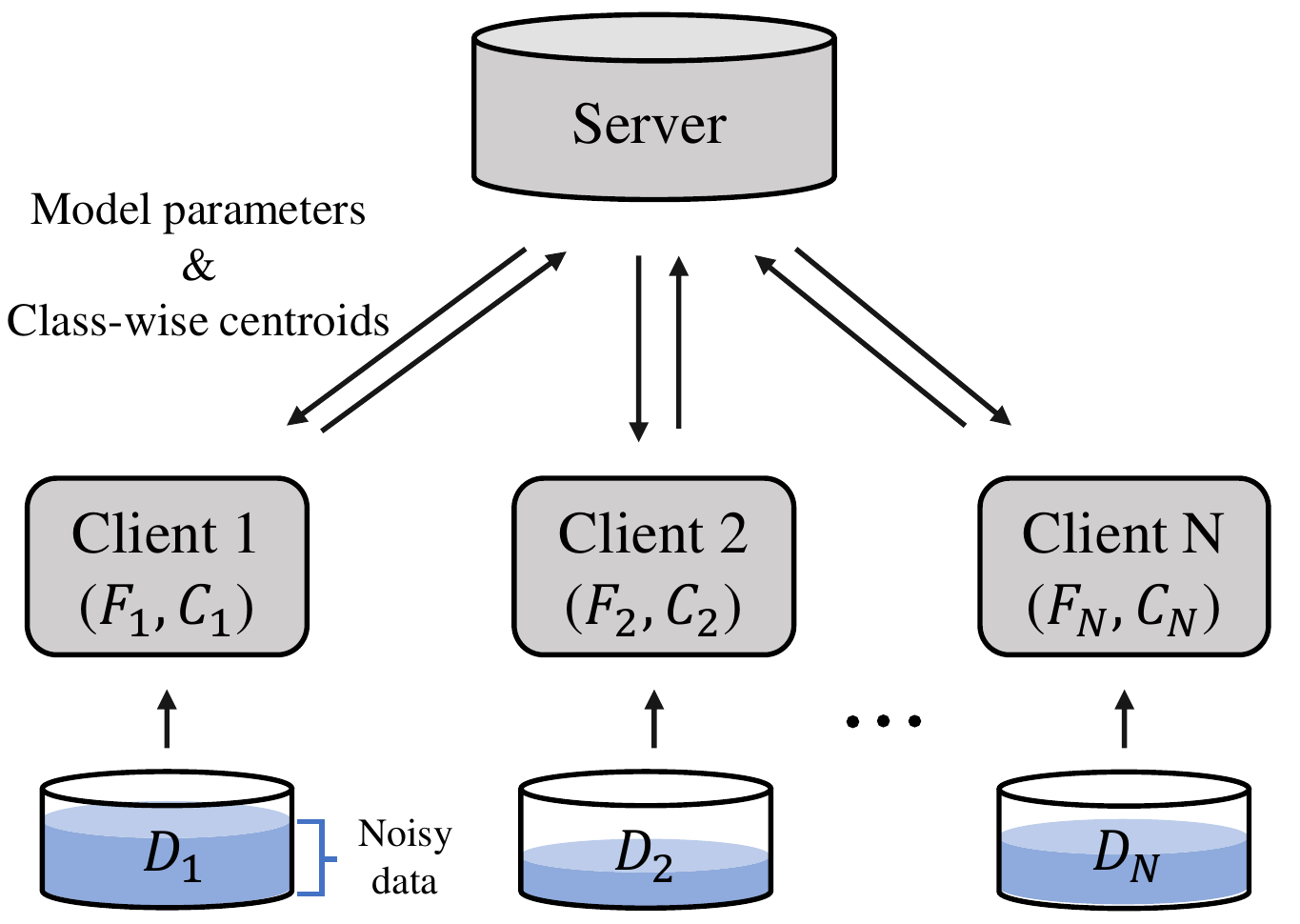,width=1\linewidth}
\centering{(a)}
\end{minipage}\hfill
\begin{minipage}{0.455\linewidth}
\epsfig{file=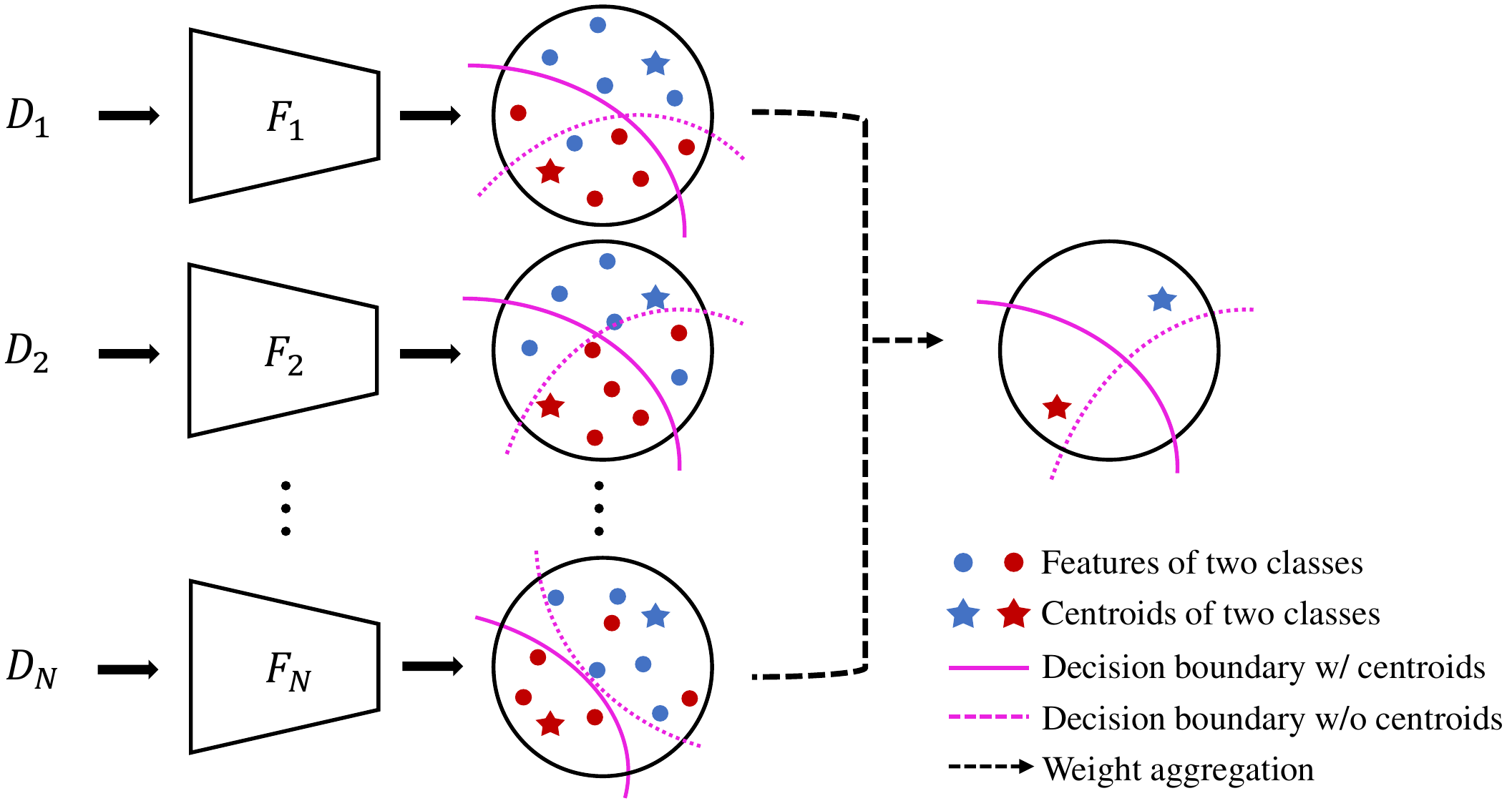,width=1\linewidth}
\centering{(b)}
\end{minipage}
\caption{(a) An illustration of the federated setting with noisy labels, where each client has a dataset with a different noise ratio. Our proposed method permits the server and the clients to interchange class-wise centroids. (b) Without any restrictions on the local model, the clients' decision boundaries can be significantly different with each other since the client's model is trained on the individual noisy dataset for a large number of local epochs. Aggregating these local models can induce the server model to have incorrect boundaries. We leverage class-wise centroids to achieve local decision boundaries similar to the others.}
\label{fig222}
\end{center}
\end{figure*}

Therefore, in federated learning with noisy labels, different noise distributions in clients should be considered, and the learning directions of clients' models should be kept similar. 
To treat these difficulties, we introduce a new federated learning scheme that the server cooperates with local models to maintain consistent decision boundaries by interchanging class-wise centroids, as described in Fig. \ref{fig222}.
In detail, we store local class-wise centroids on each device, which are central features of local data, and upload them on the server in every round. The server aggregates them into global centroids and broadcasts these centroids to clients. The centroids are used to update local models to maintain consistent decision boundaries with other clients, although the noise distributions in clients' data are different from each other.

In local updates, we compute local centroids based on samples with relatively small-losses to reduce the effect of noisy data, motivated by \cite{han2018co}.
We adjust these centroids based on the similarity with global centroids to prevent them from being corrupted by representations of noisy data.
Based on the centroids, we select confident samples to prohibit the model from fitting to noisy labels.
We also utilize a global model for unconfident samples to correct the given labels, which alleviate overfitting to noisy samples in each local model.

To the best of our knowledge, this is the first federated learning algorithm dealing with noisy labels. 
We present a new federated learning scheme interchanging additional information called centroids and propose novel algorithms for reducing the effect of noisy data.
Our approach maintains high performance on various noise ratios in the federated setting (Fig. \ref{fig1}).


\section{Related work}

\subsection{Federated learning}
Federated learning has drawn striking attention in recent years because of the increasing number of edge devices and local data collected by them. Federated learning aims to fully utilize the information of each local data without causing any serious data privacy and communication issues by transmitting the local network's parameters instead of local raw data. For preventing the server from those issues, there are several restrictions on federated learning, and they raise various problems: 1) statistical challenges (non-i.i.d. data), 2) lower network bandwidth, 3) inconsistent accuracy across devices, and 4) noisy communication. FedProx \cite{li2018federated}, FedMA \cite{Wang2020Federated}, and research about the convergence of FedAvg \cite{zhao2018federated, Li2020On} focus on the algorithms that converge the model in non-i.i.d. data. For the limitation of network bandwidth, DGC \cite{lin2018deep}, signSGD \cite{bernstein2018signsgd}, and STC \cite{sattler2019robust} only transmit important gradient changes. To maintain uniform accuracy of clients, Li et al. \cite{Li2020Fair} propose fair resource allocation. For noisy communication, Ang et al. \cite{9026922} aim to cope with the disturbance of noisy communication.
Furthermore, various studies on specific tasks considering privacy-preserving are increased in a few years, {\it e.g.}, domain adaptation \cite{Peng2020Federated}, and continual learning \cite{shoham2019overcoming, yoon2020federated}.

The aforementioned studies assume that every client has a clean dataset. However, correct annotation is not guaranteed since the local data are created by clients. Therefore, we consider that the local dataset consists of data with noisy labels and propose an algorithm to deal with it.

\subsection{Learning on noisy data}
There are many studies on the robustness of networks against noisy labels in the centralized setting. Since deep networks have sufficient capacity to fit on the whole noisy dataset \cite{zhang2016understanding}, it is essential to develop robust training methods against noisy labels. Noise cleaning-based approaches \cite{han2018co, guo2018curriculumnet, huang2019o2u, lyu2019curriculum, wei2020combating} aim to detect noisy samples and train with clean samples. In particular, Co-teaching \cite{han2018co} introduces a pair of networks with the same structure, and each network guides its peer network by using its small-loss instances. Among the label correction approaches \cite{tanaka2018joint, yi2019probabilistic}, Joint Optimization \cite{tanaka2018joint} updates all dataset labels with pseudo-labels to prevent the network from fitting onto the noisy dataset. A new type of recent research focuses on label correction by adopting the representation power of the network to distinguish clean labels \cite{lee2018cleannet, han2019deep}. Deep self-learning \cite{han2019deep} determines the label of the sample by comparing its features with several prototypes of the categories. Furthermore, meta-learning based methods \cite{ren2018learning, Li_2019_CVPR, shu2019meta} focus on optimizing parameters that are less prone to overfitting, and another effective approach is to design robust loss functions \cite{zhang2018generalized, wang2019symmetric} for a noise-tolerant model.

Previous algorithms for noisy labels aim to train networks in the centralized setting, not the federated setting. In the noisy federated setting case, clients have data with various noise distributions, and this can result in inconsistent decision boundaries and severe weight divergence in local models. To tackle these problems, we let the server cooperate with the clients to maintain consistent decision boundaries via class-wise centroids.
By exploiting the centroids, we ensure that all local models have similar feature representations of classes. Moreover, we propose an algorithm for selecting confident samples and a self-training scheme suitable for the federated setting.

\begin{figure*}
\begin{center}
\begin{minipage}{0.8\linewidth}
\centering{\epsfig{file=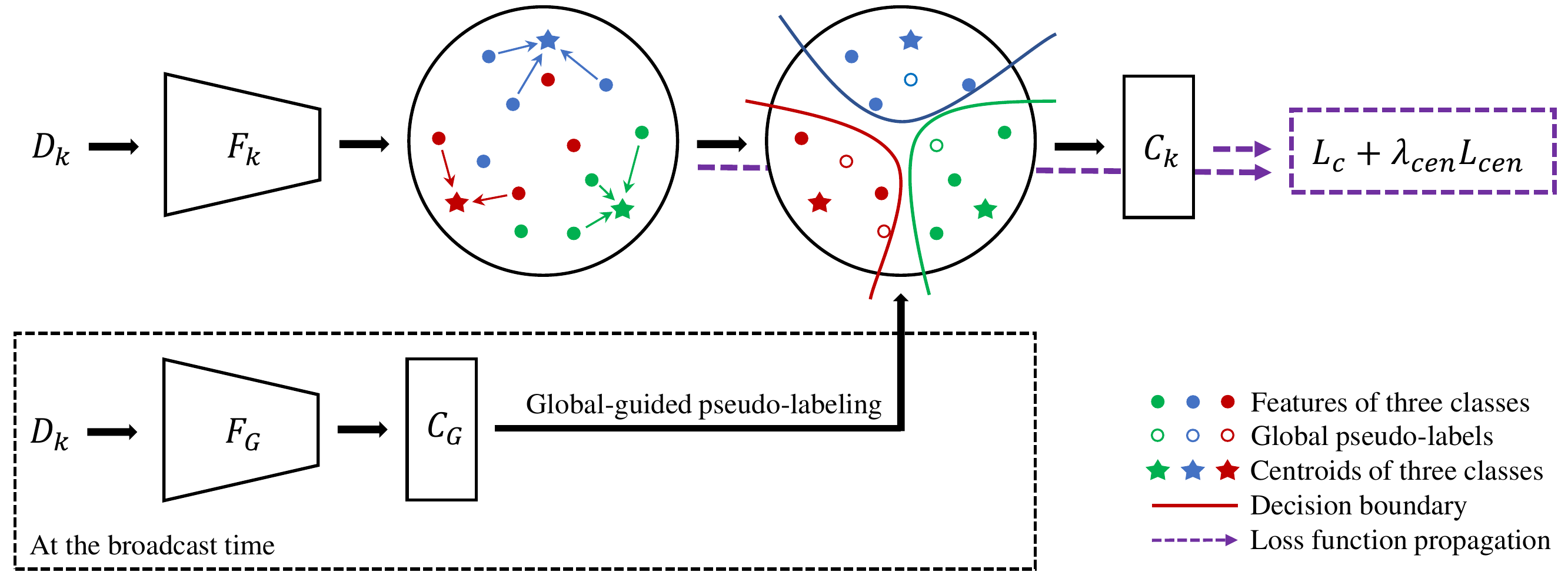,width=1.0\linewidth}}
\end{minipage}
\end{center}
   \caption{Our proposed local update algorithm. At the broadcast time, the server weights and global class-wise centroids are transmitted to each client. The client utilizes the server parameters ($F_{G}$ and $C_{G}$) for global-guided pseudo-labeling and constrains local feature representations with the global centroids.}
\label{fig3}
\end{figure*}

\section{Robust federated learning with noisy labels}
In this section, we start with the problem definition, then describe our proposed local update and global update methods.

\subsection{Problem definition and notations}
In the federated setting with multiple clients and a global server, local training data of the {\it k}-th client consist of images and the corresponding labels ${\it D}_{k} = \{(x_{k}^{i}, y_{k}^{i})\}_{i=1}^{n_{k}}$, and the server cannot access any training data.
In the noisy federated learning scenario, local datasets inevitably contain noise samples, where some of the given labels are not accurate, and noise distributions in clients' data are different from each other. The models can overfit to noisy data and suffer from weight divergence leading to aggregation difficulties in federated learning \cite{zhao2018federated, lim2020federated}.

To solve the above problem, we introduce global and local class-wise centroids, which are central features of each class in a server and clients, respectively.
Local centroids are the average feature vectors from the global average pooling layer in each local dataset, and global centroids are calculated by reflecting the local centroids of selected clients, which are depicted in the next sections in detail.
We denote global centroids and local centroids of the {\it k}-th client corresponding to class $c$ by ${\bf f}_{G}^{c}$ and ${\bf f}_{k}^{c}$. In addition, ${\bf y}_{k}^{i}$ and ${\bf \hat{y}}_{k}^{i}$ indicate the one-hot vector of the ground truth label and a pseudo-label extracted by the softmax layer, respectively.

\subsection{Local updates}
\label{local}
At the beginning of each round, selected clients receive both global model parameters and global class-wise centroids from the server for local updates. 

Before local updates, selected clients download the global model parameters, and their models are trained with their own local dataset by exploiting the following loss function:
\begin{equation}
\begin{split}
\label{c_loss}
L_{c}^{k} & = {\bf m}_{k}l_{ce}(C_{k}(F_{k}({\bf x}_{k})), {\bf y}_{k}) \\
& + ({\bf 1} - {\bf m}_{k})l_{ce}(C_{k}(F_{k}({\bf x}_{k})), {\bf \hat{y}}_{k}),
\end{split}
\end{equation}
where $F_{k}$ and $C_{k}$ denote the feature extractor and the classifier of the {\it k}-th client, respectively, and $l_{ce}(\cdot)$ is the cross-entropy loss function. A binary mask vector of the {\it k}-th client ${\bf m}_{k} \in \{0, 1\}^{n_{k}}$ controls whether it learns the ground truth label or the pseudo-label. We propose a novel sampling approach to select confident samples that are used to update the mask ${\bf m}_{k}$. In addition, we introduce a global-guided pseudo-labeling method, which takes advantage of the federated setting. Instead of a naive pseudo-labeling \cite{tanaka2018joint}, we obtain ${\bf \hat{y}}_{k}$ by exploiting the global model $F_{G}$ and $C_{G}$. This method improves local model performance while preventing the model from overfitting to noisy data.

In parallel, each client loads global centroids from the server and updates its model to have similar features with the global centroids.
To achieve this, we calculate local centroids on each local model depending on the similarity with global centroids, and we explicitly constrain local features to map local centroids.
Note that the server and all of the clients store centroids in their own devices and transmit them in each communication round. 
There is an additional communication burden required for class-wise centroids, but the amount is much smaller than model parameters (0.01\% to 0.03\%  in our experiments).

\paragraph{Local centroids.}
We use feature vectors extracted from $F_{k}$ to compute local class-wise centroids ${\bf f}_{k}$.
If we calculate ${\bf f}_{k}$ by using all local samples with given labels, noise labels have a negative effect on the correct formulation of centroids. Therefore, we introduce loss-based local centroids, which are motivated by \cite{han2018co, arpit2017closer}. We only use features of samples with relatively small-losses to create accurate feature centroids.
At first, we refine the dataset $D_{k}$ by selecting $R(t)$ percentage of small-loss instances on each client as follows:
\begin{equation}
\label{eq_small_loss}
\hat{{\it D}}_{k} = \text{argmin}_{D'_{k}: \left|{D'_{k}}\right|\geq R(t)\left|D_{k}\right|}{l_{ce}(D'_{k}}),
\end{equation}
where $D'_{k}$ is the optimization variable, $\left| \cdot \right|$ stands for the cardinality of a set the number of samples, and $R(t)$ controls how many small-loss samples should be selected in each round. 
Then, the $k$-th local model calculates naive average features of each class ${\bf \hat{f}}_{k}^{c}$ depending on the small-loss samples as follows:
\begin{equation}
\label{eq2}
{\bf \hat{f}}_{k}^{c} = \frac{1}{\tilde{n}_{k}^{c}}\sum_{x_{k}^{i}\in \hat{{\it D}}_{k}}F_{k}(x_{k}^{i})\mathds{1}(y_{k}^{i} = c),
\end{equation}
where $\tilde{n}_{k}^{c}$ is the number of samples corresponding to the label $c$ in $\hat{{\it D}}_{k}$, and $\mathds{1}(\cdot)$ is the indicator function returning 1 for true statements and 0 otherwise.

However, these average features may differ from the other clients'. To avoid these undesirable deviations, we derive local centroids ${\bf f}_{k}^{c}$ by weighted average depending on the similarity between global centroids ${\bf f}_G^{c}$ and the average features ${\bf \hat{f}}_{k}^{c}$ as follows:
\begin{equation}
\label{eq4}
{\bf f}_{k}^{c} = (1 - {sim}({\bf f}_G^{c}, {\bf \hat{f}}_{k}^{c})^2){\bf f}_G^{c} + {sim}({\bf f}_G^{c}, {\bf \hat{f}}_{k}^{c})^{2}{\bf \hat{f}}_{k}^{c},
\end{equation}
where ${sim}(\cdot,\cdot)$ can be any similarity function, but we choose cosine similarity for our experiments. ${\bf \hat{f}}_{k}^{c}$ is calculated by Eq. \ref{eq2}, and global centroids ${\bf f}_G^{c}$ are transmitted from the server reflecting entire clients' centroids, which is described in the next section. 

We expect that class-wise centroids are the central features of clean samples. At the beginning of training, deep networks tend to prioritize learning simple patterns first \cite{arpit2017closer}, and we exploit this property to form global centroids less susceptible to noisy samples. After that, we update local centroids to reflect similarity with these global centroids. This similarity-based update can keep centroids less corrupted by noisy data even after a large number of training epochs.

We exploit these local centroids to reduce weight divergence of clients' models. In detail, we design a loss function to map the features of the confident sample onto the centroids corresponding to the class as follows:
\begin{equation}
\label{eq5}
L_{cen}^{k} = \sum_{i=1}^{n_{k}}m_{k}^{i} \left \| F_{k}(x_{k}^{i}) - {\bf f}_{k}^{y_{k}^{i}} \right \|_{2}^{2},
\end{equation}
where $m_{k}^{i}$ denotes a binary mask of the {\it k}-th client that returns 1 for confident samples and 0 otherwise.

\paragraph{Confident samples.}
We introduce a sampling approach to select confident samples for training each client's model without a detrimental influence from noisy labels.
To this end, we introduce the feature similarity-based labels as follows:
\begin{equation}
\tilde{y}_{k}^{i} = \text{argmax}_{{y}}{sim}({\bf f}_{k}^{y}, F_{k}(x_{k}^{i})).
\end{equation}
Note that we should not fully trust given labels because they may not be annotated accurately. Also they should not depend on feature similarity-based labels inducing the wrong labels for the hard samples.
The complementary use of feature similarity-based and ground truth labels can help to find accurate confident samples. Therefore, we consider the similarity-based labels with local centroids and the ground truth labels at the same time.
By adopting the ground truth label and the similarity-based label together, $m_{k}^{i}$ for masking a confident sample is obtained as follows:
\begin{equation}
\label{mask}
m_{k}^{i} = \mathds{1}(\tilde{y}_{k}^{i} = y_{k}^{i}).
\end{equation}
We exploit this mask for $L_{c}$ and $L_{cen}$ to reduce the impact of noise samples. Since the number of confident samples is not fixed for each class, this mask can choose confident samples well regardless of different noise ratio for each class.

\paragraph{Global-guided pseudo-labeling.}
To fully utilize the local data information, we exploit the well-known label correction method \cite{tanaka2018joint}. 
Although this self-learning strategy with pseudo-labeling is powerful for label correction in the centralized setting, it leads local models to be self-biased \cite{arazo2019pseudo}. 
Therefore, we propose global-guided pseudo-labeling, which corrects labels of local data by employing the server model.
Our technique for the label estimation prevents local models from being self-biased. Each client receives the global model at the broadcast time and uses the model to generate global-guided pseudo-labels ${\bf \hat{y}}_{k}$ as follows:
\begin{equation}
\label{global_guided}
{\bf \hat{y}}_{k} = C_{G}(F_G({\bf x_k})),
\end{equation}
where $C_{G}$ and $F_G$ are the client's networks with global parameters.
After that, each client trains its network with these global-guided pseudo-labels by Eq. \ref{c_loss}.

Finally, the {\it k}-th local model is trained to minimize the sum of three losses:
\begin{equation}
\label{total_loss}
L_{total}^{k} = L_{c}^{k} + \lambda_{cen}L_{cen}^{k} + \lambda_{e}L_{e}^{k},
\end{equation}
where $L_{e}^{k}$ is the entropy regularization of prediction results.
Note that this term forces probability distribution of each softmax output to a single class. ${\bf p}^{i}$ indicates softmax outputs $C_{k}(F_{k}(x_{k}^{i}))$, and the loss $L_{e}^{k}$ is calculated by $-\sum _{i} {{\bf p}^{i}}\mbox{log}{{\bf p}^{i}}$.
$\lambda_{cen}$ and $\lambda_{e}$ indicate trade-off parameters. Our complete algorithm is illustrated in Fig. \ref{fig3}.

\subsection{Global updates}
\label{global}
After the local update in each round, the clients upload model parameters and local centroids to the server.
We exploit FedAvg \cite{mcmahan2016communication} for weight aggregation, which is well known as an effective algorithm for i.i.d. data.
For centroid aggregation, the server updates global centroids by a similarity-based aggregation of uploaded local centroids. 
This leads the server model to explicitly deal with the different noise ratios in clients. Moreover, since it performs a class-wise summation of local centroids, it is less affected by different noise ratios in classes.

\paragraph{Weight aggregation.}
We execute FedAvg \cite{mcmahan2016communication} for weight aggregation, which is suitable for an i.i.d. dataset.
Since only noisy labels are added in i.i.d. data, we expect that the FedAvg algorithm works well enough in our experimental settings. FedAvg takes a weighted average of local parameters $\theta_{L}$ as follows:
\begin{equation}
\label{weight_agg}
\theta_{G} = \sum _{k \in K} \frac{n_{k}}{n}\theta_{L, k},
\end{equation}
where $\theta_{G}$ is global parameters, and $n$ and $n_k$ indicate the total number of data and the number of the {\it k}-th client's data, respectively.

\paragraph{Global centroid aggregation.}
To tackle the different noise distributions in clients explicitly, we adjust global centroids to employ the similarity-based summation of local centroids. In every round, local centroids of selected clients update global centroids by using the similarity to previous global centroids in the server. Let $K$ be the set of indices of clients selected in the current round, then global centroids are updated as follows:
\begin{equation}
\label{global_centroid_agg}
{\bf f}_{G}^{c} = \frac{1}{\sum _{k\in K}w_{k}^{c}}\sum _{k \in K}w_{k}^{c}{\bf f}_{k}^{c},
\end{equation}
where $w_{k}^{c}$ indicates similarity between stored global centroids ${\bf \hat{f}}_{G}^{c}$ and the uploaded the {\it k}-th client centroids of class $c$, and it is obtained by:
\begin{equation}
w_{k}^{c} = {sim}({\bf \hat{f}}_{G}^{c}, {\bf f}_{k}^{c}).
\end{equation}
Therefore, this weight update rule allows global centroids to reflect the similarity with local centroids, which depends on the client and class. The complete pseudo-code is shown in supplementary material.

\begin{table*}
\caption{Test accuracy on the CIFAR-10 dataset with symmetric and pair flipping noise. We report the average accuracy over the last 10 rounds.}
\vspace{-1mm}
\resizebox{\linewidth}{!}{
\begin{tabular}{c|lcccccc|ccccc}
\hline
Method             & \multicolumn{12}{c}{Test Accuracy (\%)}                                                      \\ \hline \hline
Noise type         & \multicolumn{7}{c|}{Symmetric flipping}               & \multicolumn{5}{c}{Pair flipping}    \\
Noise ratio        & 0     & 0.1   & 0.2   & 0.3   & 0.4   & 0.5   & 0.6   & 0.1   & 0.2   & 0.3   & 0.4   & 0.45  \\ \hline \hline
Cross Entropy Loss \cite{mcmahan2016communication} & 89.5  &  81.8 & 73.1 &  63.1 & 54.8 &  41.9  & 35.4 &   83.2  &    73.7   &   65.4  & 56.5 & 49.5      \\ \hline
Co-teaching \cite{han2018co}        & 89.8 & 88.5 & 87.6  & 86.1 & 84.4 & 81.6 & 57.7 & 88.4 & 86.3 & 73.1 & 57.5 & 55.2 \\ \hline
Joint Optimization \cite{tanaka2018joint} & 88.4 & 86.7 & 85.4 & 82.8 & 80.1 & 76.7 & 71.2  & 86.8 & 86.2 & 86.0 & 85.4 & 83.1 \\ \hline
Ours               & 91.5 & 91.2 & 90.8 & 89.6 & 88.7 & 86.4 & 83.0 & 90.9 & 90.5 & 89.8 & 89.2 & 88.8 \\ \hline
\end{tabular}}
\label{table1}
\end{table*}

\begin{table}
\centering
\caption{Test accuracy on the Clothing1M dataset in the centralized (C. L.) and federated settings (F. L.).}
\resizebox{0.95\linewidth}{!}{
\begin{tabular}{c|cc}
\hline 
Method             & \multicolumn{2}{c}{Test Acc. (\%)} \\ \hline \hline
Setting            & C. L. & F. L.         \\ \hline \hline
Cross Entropy Loss \cite{mcmahan2016communication} & 69.5               &        70.8      \\ \hline
Co-teaching \cite{han2018co}        &        71.0        &       71.8          \\ \hline
Joint Optimization \cite{tanaka2018joint} & 72.2               & 73.5            \\ \hline
Deep self-learning \cite{han2019deep} & 74.5               &      73.6         \\ \hline
Ours               &       72.5         & 76.4            \\ \hline
\end{tabular}}
\label{table2}
\end{table}

\section{Experiments}
We adopt the following federated setting from FedAvg \cite{mcmahan2016communication}. We set the number of clients to $100$, and distribute each dataset to clients in an i.i.d. fashion. We select local epoch and local mini-batch to $5$ and $50$, respectively, considering communication efficiency and memory limitations of local devices.

\subsection{Datasets}
\paragraph{CIFAR-10.}
CIFAR-10 \cite{krizhevsky2009learning} is a benchmark dataset of 10 categories, which contains 50,000 images for training and 10,000 images for testing. We replace the ground truth labels of CIFAR-10 with two types of noisy labels: symmetric flipping \cite{van2015learning} and pair flipping \cite{han2018co}, which are described in supplementary material. Since our paper mainly focuses on the robustness in federated learning with various noisy ratios, the noise ratio $\epsilon$ is chosen from $\{0.1, 0.2, 0.3, 0.4, 0.5, 0.6\}$ for symmetric flipping and $\{0.1, 0.2, 0.3, 0.4, 0.45\}$ for pair flipping. In detail, we give noisy labels to the entire dataset with the designated noise ratio, and then randomly distribute it to 100 clients. This process induces different noise distributions in clients, and we fix the seed for a fair comparison. In ablation studies, we experiment with extremely different noise ratios.

\paragraph{Clothing1M.}
Clothing1M \cite{xiao2015learning} is a large real-world dataset of 14 categories, which contains 1 million images of clothing with noisy labels since it is obtained from several online shopping websites. In \cite{xiao2015learning}, it is reported that the overall noise ratio is approximately 38.46\%. The Clothing1M dataset also contains 50k, 14k, and 10k of clean data for training, validation, and testing, respectively, but we do not use the clean training data. For federated learning, we randomly divide the clothing1M dataset into 100 groups, which indicates the number of clients, and we set them as local datasets.

\subsection{Implementation details}
We experiment with our proposed method, Cross Entropy Loss \cite{mcmahan2016communication}, Co-teaching \cite{han2018co}, Joint Optimization \cite{tanaka2018joint}, and Deep self-learning \cite{han2019deep} in our federated setting\footnote{We use official codes for Co-teaching and Joint Optimization, and reproduce the code for Deep self-learning according to the paper.}. Note that we experiment these algorithms with FedAvg \cite{mcmahan2016communication} for weight aggregation. 
For CIFAR-10, we implemented the 9-Layer CNN applied in Co-teaching \cite{han2018co}. 
We do not report Deep self-learning because the 9-Layer CNN is not a pre-trained model, which does not have enough representation power to extract valid prototypes.
Training on the Clothing1M dataset, we exploited ResNet-50 \cite{he2016deep} pre-trained on ImageNet \cite{deng2009imagenet} by following \cite{tanaka2018joint, han2019deep}.
To prevent overfitting to a small number of training data in each local dataset, we augment training data by resizing, normalizing, and cropping images.
The detailed experimental setup is described in supplementary material.

\subsection{Analysis}
\paragraph{CIFAR-10.}
We report the results of CIFAR-10 with symmetric flipping and pair flipping in Table \ref{table1}. As shown in Table \ref{table1}, our method achieves better overall test accuracy at various noise ratios.

Co-teaching selects a fixed number of loss-based samples, which is vulnerable to different noise distributions in clients. It causes serious performance degradation of the server model since each client model is affected by noisy data.
Joint Optimization follows a self-training scheme, which is a naive pseudo-labeling method. In federated learning, this self-training method may lead the network to be self-biased due to a large number of local epochs, and the weights of self-biased local models can be diverged severely.
Notably in extremely noisy cases, clients cannot be trained properly due to high noise ratios, and when these local parameters are aggregated in the server, performance is further deteriorated.
Our proposed method is also dependent on a loss-based algorithm but robust to different noise distributions in clients because of the similarity-based centroids update. Moreover, we exploit a global-guided pseudo-labeling method, which mitigates self-bias of each client, and we validate the effectiveness of this algorithm in ablation studies.

Furthermore, neither of the two previous methods can guarantee that all local models are trained to form similar decision boundaries, which makes aggregation of local models unsuccessful.
Our proposed method constrains class representations of all client models not to diverge from global class representations.
This induces all local models to be trained to have similar boundaries, and we demonstrate the efficacy of the algorithm in noisy federated learning.

\begin{table}
\caption{Test accuracy with different noise ratios in clients (left) and classes (right) on the CIFAR-10 dataset.}
\centering
\resizebox{0.72\linewidth}{!}{
\begin{tabular}{c|c}
\hline
{$\eta$} & {Test Acc. (\%)} \\ \hline \hline
{$\epsilon$} & {0.4} \\ \hline \hline
{0}                    &               88.7                  \\ \hline
{0.1}               &                    88.5                 \\ \hline
{0.2}               &                   88.7                  \\ \hline
{0.3}               &                       88.7              \\ \hline
{0.4}                 &                 89.0               \\ \hline
\end{tabular}
\hfill \hspace{2.3mm}
\resizebox{0.405\linewidth}{!}{
\begin{tabular}{c|c}
\hline
{$\epsilon$} & {Test Acc. (\%)} \\ \hline \hline
{0.1}                    &     91.1                   \\ \hline
{0.2}               &         90.5                        \\ \hline
{0.3}               &         89.6                       \\ \hline
{0.4}               &         88.6                            \\ \hline
{0.5}                 &        86.8                    \\ \hline
{0.6}                 &        83.1                  \\ \hline
\end{tabular}}}
\label{table3}
\end{table}

\paragraph{Clothing1M.}
Different from the artificial noise in CIFAR-10, Clothing1M is a real-world noisy label dataset, including lots of unknown structure noise. By comparing the results in Table \ref{table2}, we can see that our proposed method outperforms the others by a large margin in the federated setting.
In the case of Deep self-learning \cite{han2019deep}, which corrects the labels of the data comparing the similarity with the several prototypes of the features, it achieves great performance improvement in the centralized setting. However, in the federated setting, this algorithm suffers from the significant performance degradation since it does not constrain local models from having similar decision boundaries.
We get the centroids with relatively small-losses and update them to be similar with global centroids. It can keep centroids less corrupted by noise data as well as achieve local decision boundaries similar to others.
Our algorithm is effective for the real-world noisy label dataset corrupted by unknown structure noise.

\begin{table}[]
\centering
\caption{Effect of our sampling and pseudo-labeling methods.}
\resizebox{0.6\linewidth}{!}{
\begin{tabular}{c|c|c}
\hline
mask & pseudo-label  & Acc. (\%) \\ \hline \hline
x    & x             &    74.5    \\ \hline
o    & x             &   77.8   \\ \hline
x    & naive         &   74.7     \\ \hline
o    & naive         &    86.7        \\ \hline
o    & global-guided &   88.7  \\ \hline
\end{tabular}}
\label{table22}
\end{table}

\begin{table}[]
\centering
\caption{Noisy label detection.}
\resizebox{0.97\linewidth}{!}{
\begin{tabular}{c|lcccccc}
\hline
Noise type             & \multicolumn{6}{c}{Symmetric flipping}             \\
Noise ratio        & \hspace{2mm}0.1   & 0.2   & 0.3   & 0.4   & 0.5   & 0.6 \\ \hline \hline
Precision  & 0.997 & 0.994 & 0.989 & 0.980 & 0.968 & 0.944 \\ \hline
Recall     & 0.933 & 0.937 & 0.909 & 0.903 & 0.892 & 0.840 \\ \hline
\end{tabular}
\label{table11}}
\end{table}

\subsection{Ablation studies}
We conduct ablation studies to show that each proposed algorithm is effective in federated learning with noisy labels. 

\paragraph{Different noise ratios in clients.}
In the federated setting, clients may have different amounts of noise because of the discrepancy between clients' labeling systems. In detail, we split clients into five groups and assign different noise ratios. We set noise variance $\eta$, then divide noise range $[\epsilon-\eta, \epsilon+\eta]$ equally into five noise ratios and assign the noise ratio to each group. 
For example, if we set the noise ratio $\epsilon$ and noise variance $\eta$ to $0.4$ and $0.2$, respectively, the noise ratio in each group is assigned one of $\{0.2, 0.3, 0.4, 0.5, 0.6\}$. 
In Table \ref{table3}, we experiment by fixing the noise ratio to $0.4$ and changing noise variance. Our approach achieves consistent performance regardless of the different noise ratios in clients. 

\paragraph{Different noise ratios in classes.}
Due to background knowledge of the client user, the local data can have different noise ratios in classes.
We assume an extreme situation where each client has totally erroneous samples for a single class. In detail, we force clients to have wrong labels for a single random class entirely and replace the labels of other classes with noisy labels of the designated noise ratio $\epsilon$.
We show that the proposed algorithm is robust to different noise ratios in classes in Table \ref{table3}.

\paragraph{Confident samples.} 
We evaluate the effectiveness of our sampling approach in Table \ref{table22}. Note that we set the noise ratio to $0.4$ by using symmetric flipping.
As shown in Table \ref{table22}, the mask for confident samples and the pseudo-labeling method complement each other.
The network trained only with the selected samples by removing unconfident samples has better performance than the one trained with all samples, and the performance increases considerably when the unconfident samples' labels are replaced by pseudo-labels.
Moreover, we show the precision and recall of noisy label detection of our sampling approach in Table \ref{table11}.
The precision means the number of correctly detected noisy labels over the entire number of detected noisy labels and the recall means the number of correctly detected noisy labels over the entire number of noisy labels in data.
Our sampling approach selects confident samples by the complementary use of feature similarity-based and ground truth labels. It leads to high accuracy for the precision of noise labels.

\paragraph{Global-guided pseudo-labeling.}
We have conducted the experiments by removing global-guided pseudo-labeling or replacing the proposed method with naive pseudo-labeling in Table \ref{table22}. 
Although the self-learning strategy with pseudo-labeling is powerful for label correction in centralized learning, it leads local models to be self-biased to their own datasets.
Our global-guided pseudo-labeling outperforms a naive approach to prevent local models from being self-biased.

\paragraph{Interchanging class-wise centroids.}
To show the effectiveness of interchanging centroids, we experiment our algorithm while local models are updated without using global centroids.
In detail, we use the loss function Eq. \ref{eq5} by calculating local centroids without using Eq. \ref{eq4}, which cannot explicitly constrain local models to have similar boundaries. 
Figure 4 shows that global centroids help all clients to have similar feature representations, which leads to reducing weight divergence.

\begin{figure}
\begin{minipage}{0.5\linewidth}
\begin{center}
\epsfig{file=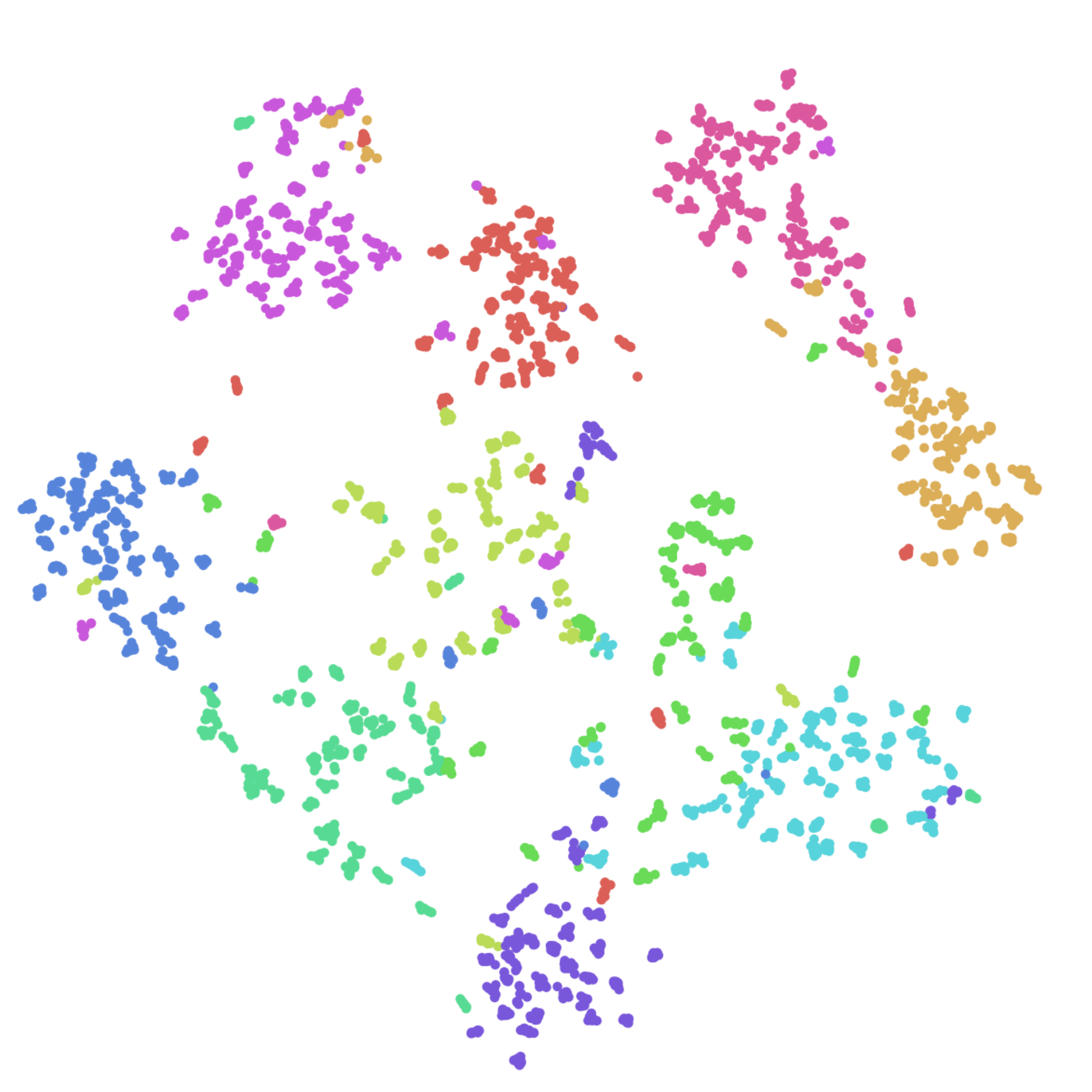,width=1\linewidth}
{(a) Ours w/o global centroids}
\end{center}
\end{minipage}\hfill
\begin{minipage}{0.5\linewidth}
\begin{center}
\epsfig{file=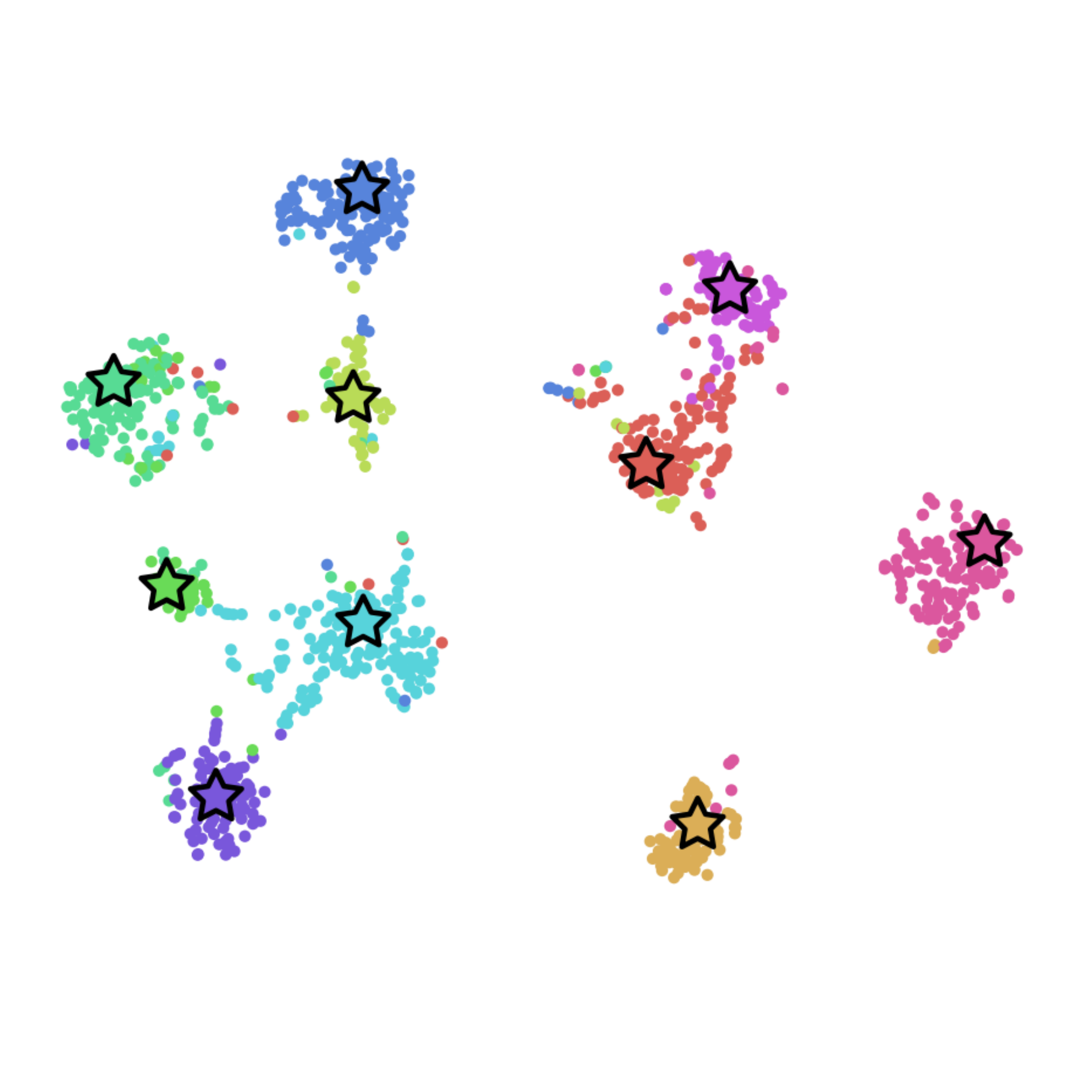,width=1\linewidth}
{(b) Ours w/ global centroids}
\end{center}
\end{minipage}
\caption{Feature visualization of selected clients with t-SNE. We plot features of clients and global centroids. (a) Data are well clustered by their categories, however varying from clients, which hinders weight aggregation in the server. (b) Most clients have similar features around global class-wise centroids marked by stars. Best viewed in color.}
\end{figure}

\section{Conclusion}
In this paper, we have considered that each local dataset may consist of noisy labels in the practical federated learning scenario. Our proposed approach is to interchange additional information, which is global and local feature centroids of classes. We demonstrate that our approach is clearly effective in the noisy federated setting by reducing weight divergence.
Moreover, we propose a novel algorithm for local updates including similarity-based confident example sampling and global-guided pseudo-labeling. In extensive experiments, we have shown that our approach outperforms existing state-of-the-art methods on CIFAR-10 and Clothing1M.

\section*{Ethics statement}
Our study suggests a practical learning scenario, especially learning with noisy labels.
Based on our proposed federated learning with noisy labels, contributors in various fields such as healthcare, fairness, and recommendation system can indirectly benefit from our global guided update scheme.
In the case of healthcare, medical data with wrong annotations can be a potential threat to the smart healthcare system.
Our scheme can help prevent a medical accident due to erroneous data from occurring in federated learning, and it would play a crucial role in the development of smart healthcare.
Our approach promotes social trends shifting a learning environment from the central server to various edge devices by allowing the models to learn without precise data.
It enables client-side learning without precise data, which does not require an expert for annotating specialized data for each device.
In our setting, we only focus on dealing with noisy labels on i.i.d. data. More work is needed for noisy labels in federated learning with non-i.i.d. data.

\small
\bibliographystyle{aaai21}
\bibliography{For_arixv}

\clearpage
\twocolumn[\section*{\LARGE{Supplementary Material: Robust Federated Learning with Noisy Labels}} \vspace{5mm}]

\section{Algorithm details}
\paragraph{Initialization of global centroids.}
Since we update local centroids using similarities with global centroids, randomly initialized global centroids hinder local models from deriving accurate local centroids.
For this reason, average features ${\bf \hat{f}}_{k}$ are used instead of global centroids in the first round. After that, global centroids are computed by aggregating clients' local centroids.
\vspace{-1mm}
\paragraph{Loss-based centroids.}
At the beginning of training, deep networks tend to prioritize learning simple patterns first \cite{arpit2017closer}. We utilize this property to keep more instances at the start, {\it i.e.}, $R(t)$ is large in Eq. \ref{eq_small_loss}. As the training proceeds, we gradually reduce $R(t)$ to prevent local models from fitting to noise samples following \cite{han2018co}. We set $T$ and $\tau$ to $10$ and $\epsilon$ in our experiments, and we show that our approach is robust to these parameters in the experimental section.
\vspace{-1mm}
\paragraph{Pseudo-labels.} In Eq. \ref{c_loss}, we train the network with unconfident samples by using pseudo-labels. At the early stage, the network cannot generate accurate pseudo-labels ${\bf \hat{y}}$ due to insufficient training time.
Therefore, at first, we replace pseudo-labels ${\bf \hat{y}}$ with ground truth labels ${\bf y}$.
After the number of rounds reaches predefined number ($T_{pl}$), we exploit pseudo-labels, then we train the network jointly by ${\bf y}$ and ${\bf \hat{y}}$.
\vspace{-1mm}
\paragraph{Scheduling for $\lambda_{cen}$.} To avoid the noisy mask at the early stage of the training procedure, we initialize $\lambda_{cen}$ to 0 and gradually increase it to a predefined number.
\vspace{-1mm}
\paragraph{Mini-batch algorithm.}
We modify Eq. \ref{eq4} for mini-batch SGD as follows:
\begin{equation}
\label{eq13}
{\bf f}_{k, j} = (1 - {sim}({\bf f}_{k, j-1}, {\bf \hat{f}}_{k, j})^2){\bf f}_{k, j-1} + {sim}({\bf f}_{k, j-1}, {\bf \hat{f}}_{k, j})^{2}{\bf \hat{f}}_{k, j},
\end{equation}
where ${\bf f}_{k, j}$ indicates local centroids at $j$-th iteration. The full algorithm for our local and global updates is given in Algorithm 1. Note that ${\bf f}_{k, 0}$ indicates global centroids ${\bf f}_{G}$ in the first local epoch.

\section{Two types of noisy labels}
Since CIFAR-10 \cite{krizhevsky2009learning} and MNIST \cite{lecun1998gradient} are clean, following \cite{reed2014training, patrini2017making}, we manually corrupt these datasets by the label transition matrix $Q$, where $Q_{ij} =$ Pr$(\tilde{y}=j | y=i)$ given that noisy $\tilde{y}$ is flipped from clean $y$.
For symmetric flipping, we inject the symmetric label noise as follows:
\begin{equation}
Q = \begin{bmatrix}
 1-\epsilon& \frac{\epsilon}{n-1} & \cdots & \frac{\epsilon}{n-1} & \frac{\epsilon}{n-1}\\ 
 \frac{\epsilon}{n-1}& 1-\epsilon &  &  & \frac{\epsilon}{n-1}\\ 
 \vdots&  & \ddots &  & \vdots\\ 
 \frac{\epsilon}{n-1}&  &  & 1-\epsilon & \frac{\epsilon}{n-1}\\ 
 \frac{\epsilon}{n-1}& \frac{\epsilon}{n-1} & \cdots & \frac{\epsilon}{n-1} & 1-\epsilon 
\end{bmatrix},
\end{equation}
where $n$ is the number of classes and $\epsilon$ indicates the noise ratio.
Pair flipping is a well-known noise generation method that focuses on fine-grained classification with noisy labels, and its noise transition matrix $Q$ is obtained as follows:
\begin{equation}
Q = \begin{bmatrix}
 1-\epsilon & \epsilon & 0 & \cdots & 0\\ 
 0& 1-\epsilon & \epsilon &  & 0\\ 
 \vdots&  & \ddots &  & \vdots\\ 
 0&  &  & 1-\epsilon & \epsilon\\ 
 \epsilon& 0 & \cdots & 0 & 1-\epsilon
\end{bmatrix}.
\end{equation}

\begin{table*}
\begin{tabular}{l}
\hline \hline
{\bf Algorithm 1}: Robust Federated Learning with Noisy Labels 
 \\ \hline
 {\bf Input}: global weights ${\bf \theta}_{G}$, global centroids ${\bf f}_{G}$, learning rate $\eta$, start round that uses pseudo-labels $T_{pl}$, fixed $\tau$,\\ round $T$ and $T_{max}$;\\
 {\bf Server executes:} \\
 \hspace{10mm}initialize ${\bf \theta}_{G}$; \\
 \hspace{10mm}{\bf for} each round $t = 1, ..., T_{max}$ {\bf do} \\
 \hspace{20mm}$S_{t}$ $\leftarrow$ (random set of m clients);\\
 \hspace{20mm}{\bf for} each client $k \in S_{t}$ {\bf in parallel do} \\
 \hspace{30mm}{\bf Load} ${\bf \theta}_{k}$, ${\bf f}_{k}$ $\leftarrow$ LocalUpdate($k$, $t$, ${\bf \theta}_{G}$, ${\bf f}_{G}$, $R(t)$);\\
\hspace{20mm}{\bf Update} global weights ${\bf \theta}_{G}$ by Eq. \ref{weight_agg};\\
\hspace{20mm}{\bf Update} global centroids ${\bf f}_{G}$ by Eq. \ref{global_centroid_agg};\\
\hspace{20mm}{\bf Update} $R(t) = 1-\text{min}\{\frac{t}{T}\tau, \tau\}$;\\
 \\
 {\bf function} LocalUpdate($k$, $t$, ${\bf \theta}_{G}$, ${\bf f}_{G}$, $R(t)$): // Run on client k\\
 \hspace{10mm}{\bf Load} ${\bf \theta}_{k} \leftarrow {\bf \theta}_{G}$;\\
 \hspace{10mm}{\bf if} $t = 1$ {\bf then} // Initialization of global centroids\\
 \hspace{20mm}{\bf Obtain} naive average features ${\bf \hat{f}}_{k}$ by Eq. \ref{eq2} from ${D}_{k}$;\\
 \hspace{20mm}{\bf Load} ${\bf f}_{k, 0} \leftarrow {\bf \hat{f}}_{k}$;\\
 \hspace{10mm}{\bf else}\\
 \hspace{20mm}{\bf Load} ${\bf f}_{k, 0} \leftarrow {\bf  f}_{G}$\\
 \hspace{10mm}{\bf Obtain} global-guided pseudo labels ${\bf \hat{y}}$ by Eq. \ref{global_guided} from ${D}_{k}$; \\
 \hspace{10mm}{\bf for} each local epoch $i = 1, ..., E$ {\bf do} \\
 \hspace{20mm}{\bf Shuffle} training set $D_{k}$; \\
 \hspace{20mm}{\bf for} $j = 1, ..., N_{max}$ {\bf do} \\
 \hspace{30mm}{\bf Fetch} mini batch $D_{k, j}$ from $D_{k}$;\\
 \hspace{30mm}{\bf Obtain} small-loss sets $\hat{D}_{k, j}$ by Eq. \ref{eq_small_loss} from $D_{k, j}$;\\
 \hspace{30mm}{\bf Obtain} confident mask vector ${\bf m}_{k, j}$ by Eq. \ref{mask};\\
 \hspace{30mm}{\bf if} $t < T_{pl}$ {\bf then}\\
 \hspace{40mm}{\bf Load} ${\bf \hat{y}} \leftarrow {\bf y}$; // Replacing pseudo-labels with ground truth labels \\
 \hspace{30mm}{\bf Update} local weights ${\bf \theta}_{k}$ by minimizing Eq. \ref{total_loss};\\
 \hspace{30mm}{\bf Obtain} loss-based average features ${\bf \hat{f}}_{k, j}$ by Eq. \ref{eq2} from $\hat{D}_{k, j}$;\\
 \hspace{30mm}{\bf Update} local centroids ${\bf f}_{k, j}$ by Eq. \ref{eq13};\\
 \hspace{20mm}{\bf Load} ${\bf f}_{k, 0}$ $\leftarrow$ ${\bf f}_{k, N_{max}}$;\\
 \hspace{10mm}{\bf Output}: ${\bf \theta}_{k}$ and  ${\bf f}_{k, 0}$.\\
 
\hline
\end{tabular}
\end{table*}

\begin{table*}[]
\caption{Test accuracy on the MNIST dataset with symmetric and pair flipping noise. We report the average accuracy over the last 10 rounds.}
\resizebox{\linewidth}{!}{
\begin{tabular}{c|lcccccc|ccccc}
\hline
Method             & \multicolumn{12}{c}{Test Accuracy (\%)}                                                      \\ \hline \hline
Noise type         & \multicolumn{7}{c|}{Symmetric flipping}               & \multicolumn{5}{c}{Pair flipping}    \\
Noise ratio        & \hspace{2mm}0     & 0.1   & 0.2   & 0.3   & 0.4   & 0.5   & 0.6   & 0.1   & 0.2   & 0.3   & 0.4   & 0.45  \\ \hline \hline
Cross Entropy Loss \cite{mcmahan2016communication} & 99.6 & 98.7 & 97.4 & 94.2 & 88.1 & 79.2 & 62.5 & 98.7 & 95.6 & 86.8 & 69.4 & 61.6\\ \hline
Co-teaching \cite{han2018co} & 99.4 & 99.4 & 99.2 & 99.1 & 98.9 & 98.1 & 96.7 & 99.3 & 99.2 & 99.1 & 96.6 & 92.3\\ \hline
Joint Optimization \cite{tanaka2018joint} & 98.1 & 97.9 & 97.9 & 97.3 & 97.0 & 97.1 & 94.1 & 98.0 & 97.3 & 97.3 & 97.3 & 97.2 \\ \hline
Ours               & 99.5 & 99.4 & 99.3 & 99.3 & 99.2 & 99.1 & 98.8 & 99.4 & 99.4 & 99.3 & 99.2 & 99.1 \\ \hline
\end{tabular}}
\label{table1_1}
\end{table*}

\section{Implementation details}
We used the Pytorch framework \cite{paszke2017automatic} to implement the model, and training was done on a GTX 1080Ti and Intel i7-6700k@4.00GHz. We utilized the official code provided by authors for Co-teaching \cite{han2018co}. We modified the official code to Pytorch version of Joint Optimization \cite{tanaka2018joint} and reproduced the model in Deep self-learning \cite{han2019deep} according to the paper.
In our federated setting, we set the number of clients to $100$, and $10$ clients are selected at every round. Batch size and local epoch are $50$ and $5$, and we trained the network during $100$, $1000$, and $40$ rounds for MNIST, CIFAR-10, and Clothing1M, respectively.
We used SGD optimizer with a momentum of 0.5, a weight decay of $10^{-4}$ for all algorithms.
Next, we describe the parameters for each algorithm.

\begin{table*}
\begin{minipage}[t]{.33\linewidth}
\captionof{figure}{Image variances in intensity.}
\vspace{1mm}
\hspace{4mm}
\includegraphics[width=23mm]{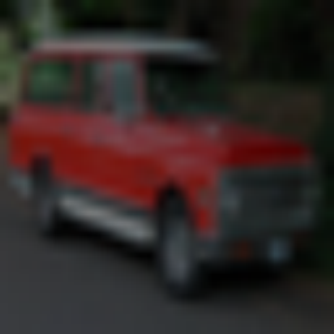}
\hspace{2mm}
\includegraphics[width=23mm]{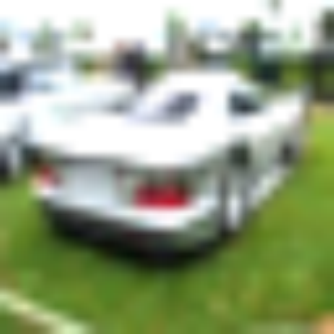}
\label{newtable3}
\end{minipage}
\vspace{-6.6mm}
\begin{minipage}[t]{.31\linewidth}
\centering
\caption{Test accuracy on the various number of participating clients in each round.}
\vspace{-1mm}
\begin{tabular}{c|ccc}
\hline
\multicolumn{1}{c|}{\# of clients} & 1  & 2  & 5   \\ \hline
\multicolumn{1}{c|}{Acc. (\%)}     & 78.6 & 85.9 & 88.2 \\ \hline \hline
\multicolumn{1}{c|}{\# of clients} & 20 & 50 & 100 \\ \hline
\multicolumn{1}{c|}{Acc. (\%)}     & 89.6 & 90.1 & 89.9 \\ \hline
\end{tabular}
\label{newtable1}
\end{minipage}
\hspace{3mm}
\begin{minipage}[t]{.32\linewidth}
\caption{Computational time. GTX 1080Ti and Intel i7-6700k@4.00GHz.}
\centering
\vspace{2.9mm}
\begin{tabular}{c|c}
\hline
Method        & time (s)  \\ \hline \hline
Co-teaching  &  45.8 \\ \hline
Joint Optimization  &  24.9 \\ \hline
Ours          &   27.8   \\ \hline
\end{tabular}
\label{newtable2}
\end{minipage}
\end{table*}

\paragraph{Ours.}
We determined balance parameters ($\lambda_{cen}$ and $\lambda_{e}$) and $T_{pl}$ based on ablation studies and the previous work \cite{tanaka2018joint}. We set $\lambda_{cen}$ and $\lambda_{e}$ to $1.0$ and $0.8$ for all datasets, and $T_{pl}$ to $30$ for MNIST, $100$ for CIFAR-10, and $5$ for Clothing1M. The initial learning rate is $0.25$ for MNIST and CIFAR-10, and $0.01$ for Clothing1M. For the Clothing1M dataset, the learning rate was decreased by 10 every 10 rounds. In addition, to obtain loss-based centroids, we set $T=10$ and $\tau = \epsilon$ by following \cite{han2018co}.
We show that our approach is robust to all these parameters in ablation studies.
For the centralized setting, we modified our algorithm by replacing global-guided pseudo-labeling with naive pseudo-labeling and calculating global centroids to use all data.

\paragraph{Co-teaching.}
We took two networks with the same architecture but different initializations as two classifiers.
Different from \cite{han2018co}, we used SGD optimizer with an initial learning rate of 0.15 for federated learning \cite{mcmahan2016communication}. We found that this setting has similar test accuracy to the original setting in the centralized setting through experiments.
We set $R(t) = 1-\tau\text{min}(t/T, 1)$ with $T = 10$ for MNIST and CIFAR-10 and $T = 5$ for Clothing1M, and $\tau = \epsilon$ by following \cite{han2018co, wei2020combating}.
We initialized the learning rate to $0.15$ for MNIST and CIFAR-10, and $0.001$ for Clothing1M. For Clothing1M, the learning rate was decreased by 10 every 10 rounds.

\paragraph{Joint Optimization.}
We determined balance parameters ($\lambda_{\alpha}$ and $\lambda_{\beta}$), start epoch, and learning rate based on \cite{tanaka2018joint}. For the MNIST and CIFAR-10 datasets, we used the different learning rates suitable for each noise ratio following \cite{tanaka2018joint}. We set $\alpha$, $\beta$, and start epoch to $1.2$, $0.8$, and $100$ respectively in MNIST and CIFAR-10.
For Clothing1M, we used a learning rate of $8 \times 10^{-4}$, and used 2.4 for $\alpha$ and 0.8 for $\beta$. 

\paragraph{Deep self-learning.}
The initial learning rate was $0.002$ and decreased by $10$ every $5$ epochs.
We followed the paper \cite{han2019deep} to choose hyper-parameters except the number of selected samples and prototypes. In the federated setting, each client only has a small portion of the original dataset. For this reason, in the label correction phase, we randomly sample 128 images for each class and $3$ class prototypes are picked out for each class.

\begin{figure*}[!t]
\begin{subfigure}{.33\textwidth}
  \centering
  \includegraphics[width=1\linewidth]{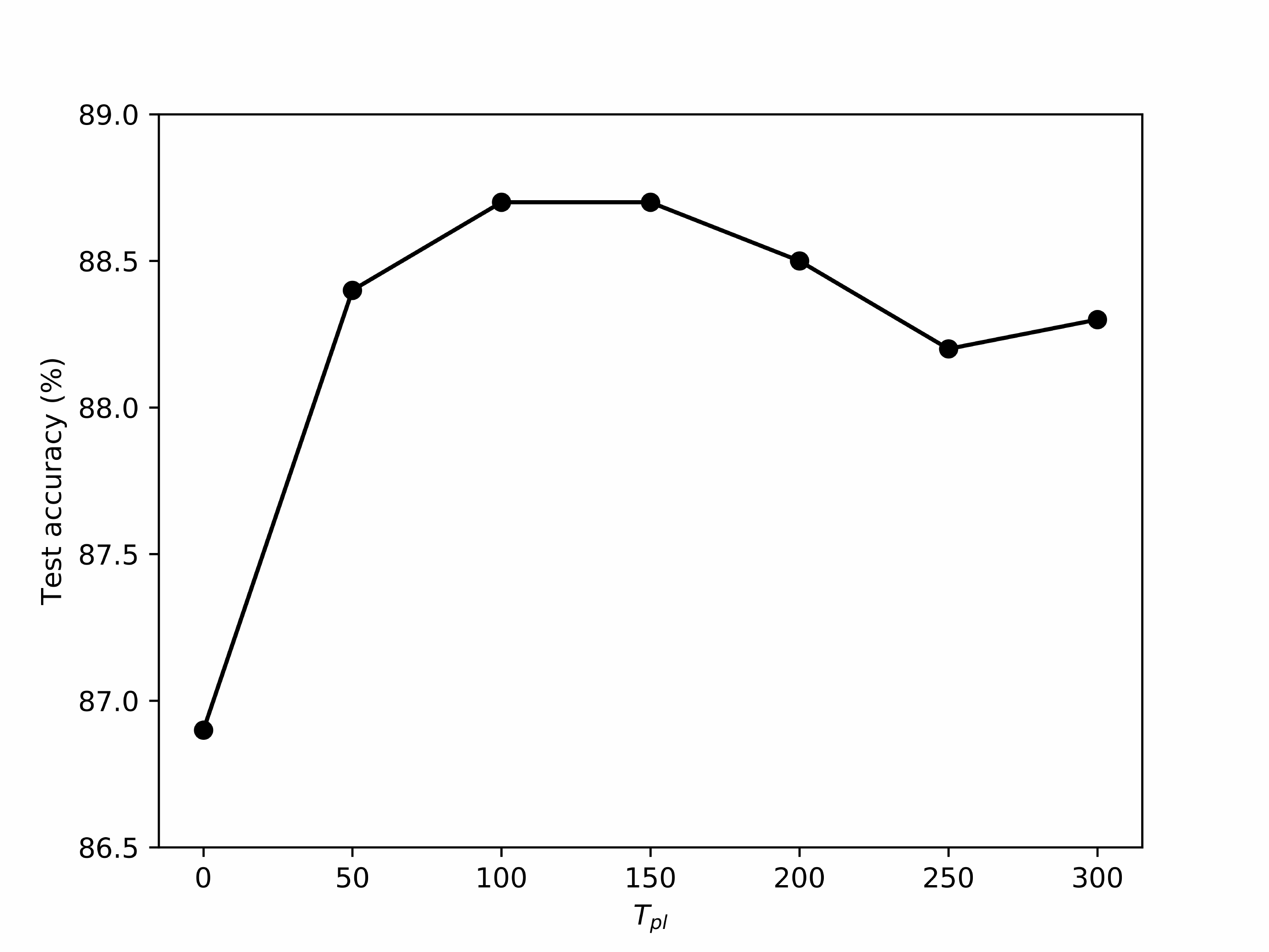}
  \caption{$T_{pl}$}
\end{subfigure}%
\begin{subfigure}{.33\textwidth}
  \centering
  \includegraphics[width=1\linewidth]{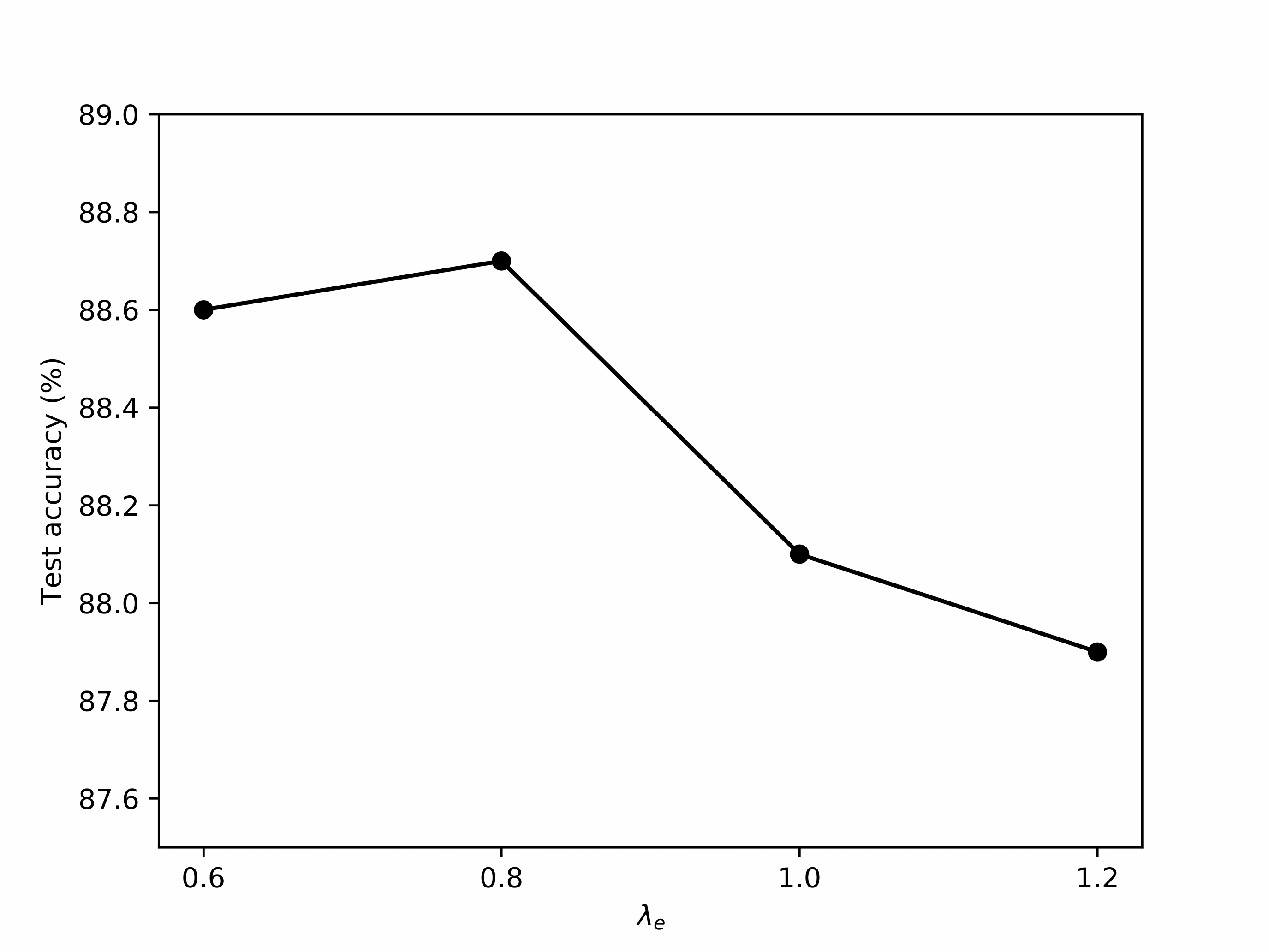}
  \caption{$\lambda_{e}$}
\end{subfigure}%
\begin{subfigure}{.33\textwidth}
  \centering
  \includegraphics[width=1\linewidth]{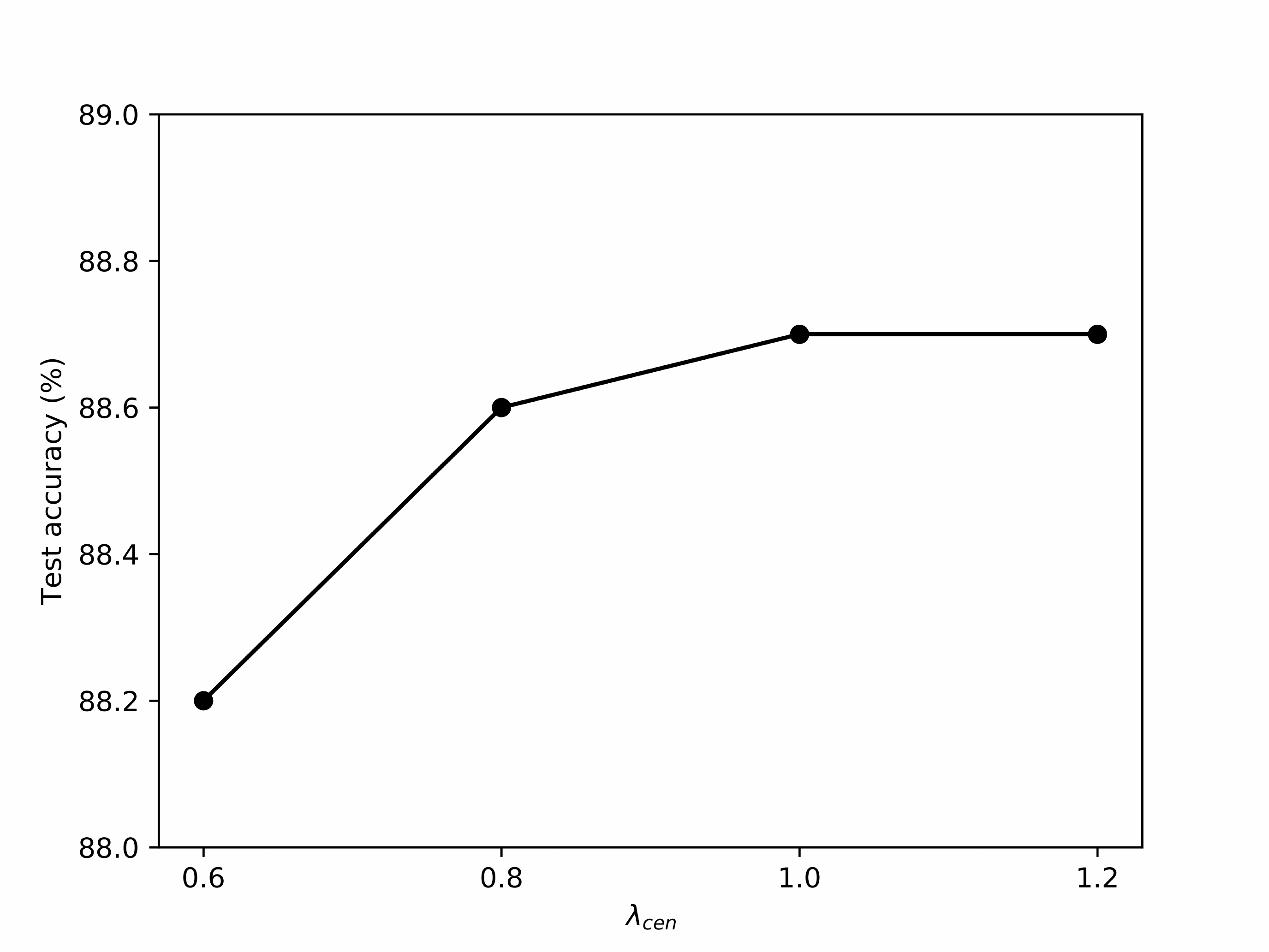}
  \caption{$\lambda_{cen}$}
\end{subfigure}%
\hfill

\begin{subfigure}{.33\textwidth}
  \centering
  \includegraphics[width=1\linewidth]{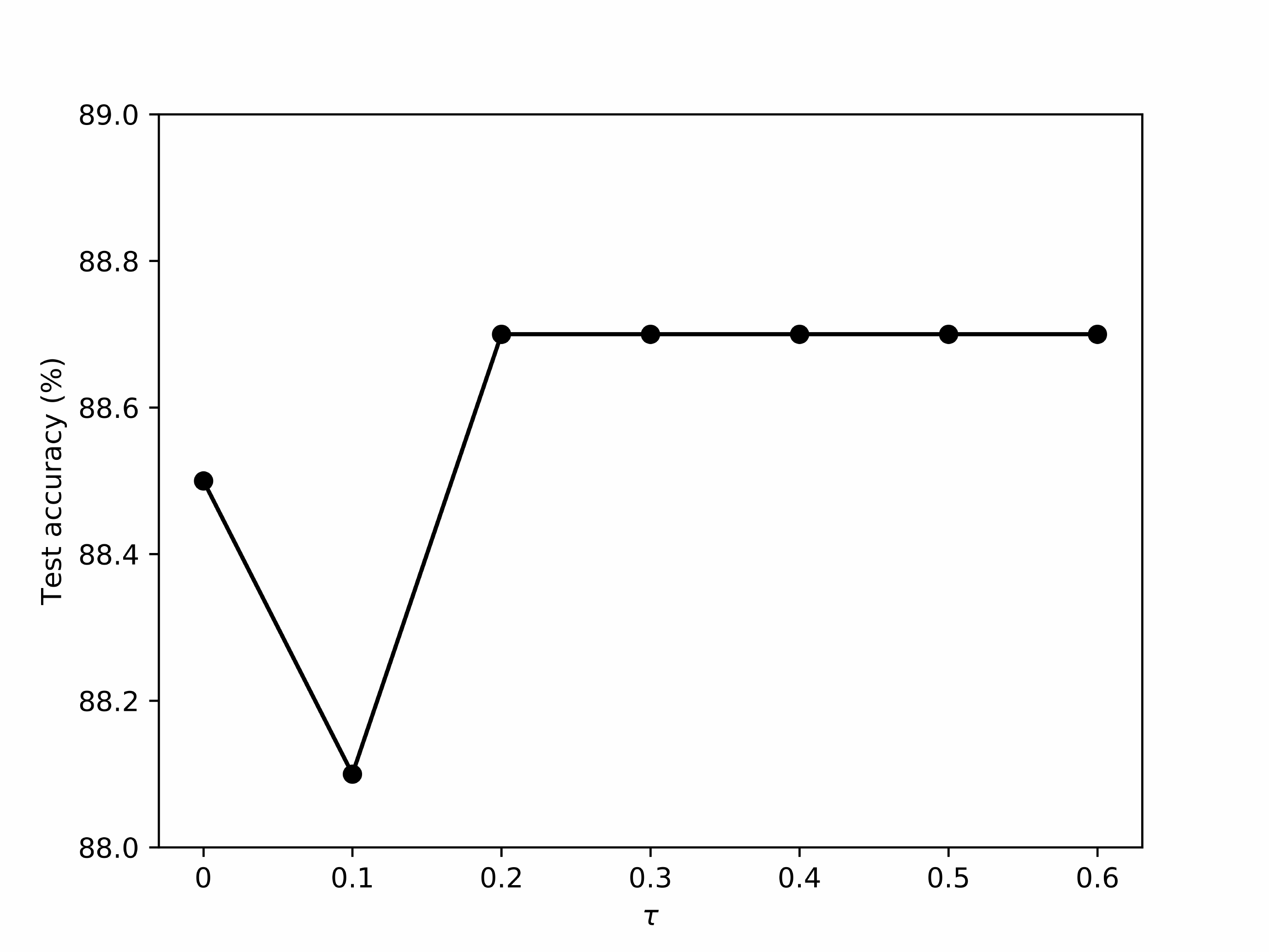}
  \caption{$\tau$}
\end{subfigure}%
\begin{subfigure}{.33\textwidth}
  \centering
  \includegraphics[width=1\linewidth]{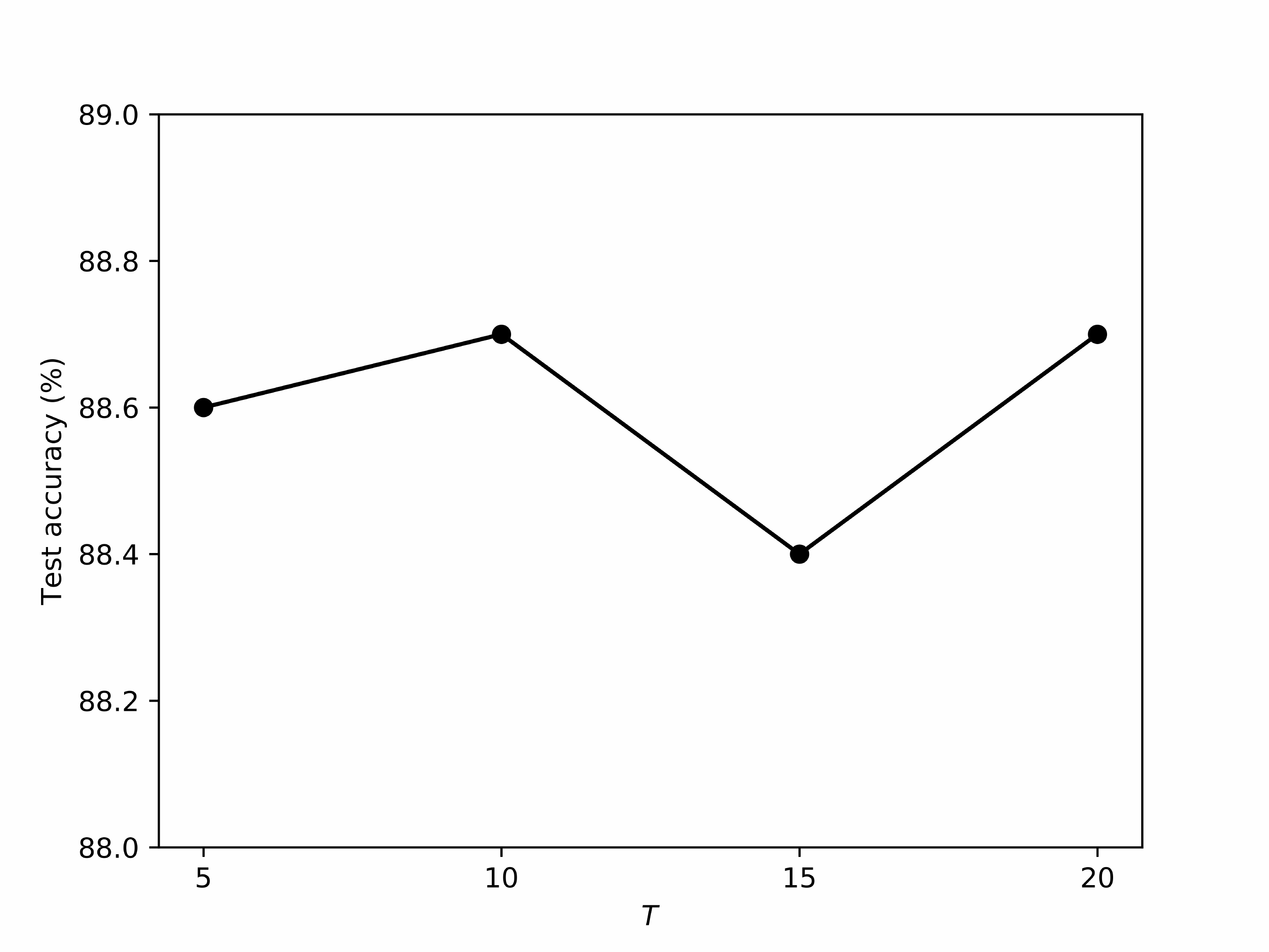}
  \caption{$T$}
\end{subfigure}%
\caption{Performance dependency of hyper-parameters.}
\label{ablation}
\end{figure*}

\section{More experimental results}

\paragraph{Experimental results on MNIST.} MNIST \cite{lecun1998gradient} is a benchmark dataset of 10 categories, which contains 60,000 images for training and 10,000 images for testing.
We replace the ground truth labels of MNIST with two types of noisy label: symmetric and pair flipping. We implemented the 9-Layer CNN applied in Co-teaching \cite{han2018co}. In Table \ref{table1_1}, our proposed approach maintain high performance at various noise ratios.

\paragraph{Image variances in different domains.}
In the real world, the data would differ from client to client not only in terms of noise in the labels but also the features themselves.
We investigate this situation in federated learning. We assume that clients are in different environments, especially light intensity, {\it e.g.}, a photo of a car in one client would be different from that in others in terms of light intensity, as illustrated in Fig. \ref{newtable3}.
In detail, we use the noisy CIFAR-10 dataset with noise ratio $0.4$, and change the light intensity of each client dataset at a rate of specific value within 0.5 to 1.5 times by using the ImageEnhance module in the PIL package.
Our algorithm achieves $88.4\%$, although the noise exists in both labels and features ($88.7\%$ in the original setting).

\paragraph{Real federated learning scenario.}
For centralized learning, transmitting medical data from each medical center to the central server causes privacy issues.
In consideration of these issues, we experiment our algorithm under the assumption that the medical dataset is in each medical center.
We choose the COVIDx Dataset \cite{wang2020covid}, which consists of 15,282 chest x-ray images, and distribute the dataset to $100$ medical centers (clients).
Our approach achieves $93.5\%$, which is comparable to the centralized setting ($93.6\%$).

\paragraph{Number of participating clients.}
In real-world federated learning scenarios, the population base can be significantly larger and a considerably smaller portion can be selected every round.
We provide experimental results on the noisy CIFAR-10 dataset with noise ratio $0.4$ by changing the number of participating clients per round in Table \ref{newtable1}. Note that the client population is $100$.
Even if only two clients participate in communication, it shows comparable performance by reducing weight divergence of clients' models.

\paragraph{Computational cost.}
Since our proposed algorithm increases computation-cost not only for local updates but also for global updates, we measure the time from the start of the round to the next round. As shown in Table \ref{newtable2}, the speed of our algorithm is similar to that of Joint Optimization \cite{tanaka2018joint}. Due to the use of similarity-based updates and confident samples, our algorithm has a marginal increase in computational cost.
Since Co-teaching \cite{han2018co} exploits two networks, its computational time is longer than others that use only one network.

\paragraph{Performance dependency of hyper-parameters.}
We use the noisy CIFAR-10 dataset and set the noise ratio $\epsilon$ to $0.4$ by using symmetric flipping.
We set $T_{pl}$, $\lambda_{e}$, $\lambda_{cen}$, $\tau$, $T$ to $100$, $0.8$, $1.0$, $0.4$, $10$, respectively.
Then, we have conducted various experiments changing hyper-parameters, {\it i. e.,} $T_{pl}$, $\lambda_{e}$, $\lambda_{cen}$, $\tau$, and $T$.
As shown in Fig. \ref{ablation}, the prediction accuracy is robust to the hyper-parameters except $T_{pl}$ that is related to replacing ground truth labels with pseudo labels.
Since the network cannot generate accurate pseudo-labels ${\bf \hat{y}}$ at the early stage, it achieves lower performance compared to networks trained with high $T_{pl}$.

\end{document}